\documentclass[10pt,twocolumn,letterpaper]{article}

\usepackage{cvpr}              
\usepackage{graphicx}
\usepackage{amsmath}
\usepackage{amssymb}
\usepackage{booktabs}
\usepackage{multirow}
\usepackage{placeins}
\usepackage{pifont}
\newcommand{\xmark}{\text{\ding{55}}}
\usepackage{makecell}

\usepackage[pagebackref,breaklinks,colorlinks]{hyperref}
\usepackage[capitalize]{cleveref}
\usepackage{lipsum}

\newcommand\blfootnote[1]{%
  \begingroup
  \renewcommand\thefootnote{}\footnote{#1}%
  \addtocounter{footnote}{-1}%
  \endgroup
}
\crefname{section}{Sec.}{Secs.}
\Crefname{section}{Section}{Sections}
\Crefname{table}{Table}{Tables}
\crefname{table}{Tab.}{Tabs.}

\def\cvprPaperID{*****} 
\def\confName{CVPR}
\def\confYear{2024}

\begin{document}

\title{PUDD: Towards Robust Multi-modal Prototype-based Deepfake Detection}
\author{Alvaro Lopez Pellicer$^*$, Yi Li$^*$, Plamen Angelov\\
School of Computing and Communications, Lancaster University\\}


\maketitle
\def\thefootnote{*}\footnotetext{These authors contributed equally to this work}
\begin{abstract}
Deepfake techniques generate highly realistic data, making it challenging for humans to discern between actual and artificially generated images. Recent advancements in deep learning-based deepfake detection methods, particularly with diffusion models, have shown remarkable progress. However, there is a growing demand for real-world applications to detect unseen individuals, deepfake techniques, and scenarios. To address this limitation, we propose a Prototype-based Unified Framework for Deepfake Detection (PUDD). PUDD offers a detection system based on similarity, comparing input data against known prototypes for video classification and identifying potential deepfakes or previously unseen classes by analyzing drops in similarity. Our extensive experiments reveal three key findings: (1) PUDD achieves an accuracy of 95.1$\%$ on Celeb-DF, outperforming state-of-the-art deepfake detection methods; (2) PUDD leverages image classification as the upstream task during training, demonstrating promising performance in both image classification and deepfake detection tasks during inference; (3) PUDD requires only 2.7 seconds for retraining on new data and emits 10$^{5}$ times less carbon compared to the state-of-the-art model, making it significantly more environmentally friendly. \blfootnote{Emails: \{a.lopezpellicer, y.li154, p.angelov\}@lancaster.ac.uk. (\textit{Corresponding author: Yi Li.})} 
\end{abstract}
\section{Introduction}
\label{sec:intro}
Deepfakes, created through digitally manipulated techniques, convincingly replace one person's likeness with another's \cite{deepfake, fake}. A recent study highlights the challenge in distinguishing real from AI-generated images, with only 61$\%$ of participants accurately identifying them—falling short of the expected 85$\%$ \cite{new1}. This difficulty stems from advancements in deep learning models, notably autoencoders \cite{aedeep, TAI}, Generative Adversarial Networks (GANs) \cite{gan}, diffusion models \cite{stablefu} and neural style transfer (NST) \cite{nst}. Manipulating audio, video, and image data poses serious consequences, including security vulnerabilities, safety concerns, ethical dilemmas, and erosion of public trust \cite{fake1}. One famous example of a deepfake involved a video purportedly showing the Ukrainian president urging soldiers to surrender to Russia. The video circulated on social media and appeared on a Ukrainian news website before being debunked and removed \cite{news}. Consequently, deepfake detection has become a critical area of research, drawing increasing attention from researchers.

In recent years, significant advancements in deep learning techniques have greatly improved their effectiveness in detecting deepfakes, resulting in notable performance gains \cite{fdetect2}. Deepfake detection methods can be broadly categorized into two aspects: artifact-specific \cite{detectc1} and undirected approaches \cite{detectc2}, depending on the data and deepfake techniques involved. For instance, artifact-specific approaches focus on detecting unnatural areas in deepfake human faces by leveraging edges and optical flow. Chintha et al. employed a combination dataset comprising visual frames, edge maps, and dense optical flow maps as inputs to a recurrent XceptionNet \cite{fdetect}. By learning a fused representation of these features, the model achieves accurate predictions. On the other hand, undirected approaches eschew specific artifacts or predefined feature sets, instead training a general-purpose classifier to autonomously analyze the entire input data and learn relevant features. However, these undirected deepfake detection methods suffer from three main drawbacks.

The majority of recent deepfake detection techniques \cite{fdetect1, fdetect2, fdetect3} struggle with robustness, which refers to the ability of the detector to maintain high accuracy when processing unseen deepfakes—those generated using techniques and models different from those used in training. Robustness is essential for the practical application of these systems in real-world scenarios. Secondly, training these deepfake detection models is time-consuming due to the large scale of the network models. For example, Zhao et al. utilize two parallel Vision Transformer-Large (ViT-L) networks with several Xception blocks \cite{vit, Xception} to extract spatial and temporal features from deepfake videos \cite{fdetect1}, resulting in over 600 million parameters. Retraining such models for new individuals is therefore exceedingly time-consuming. Thirdly, many of these detection techniques lack interpretability due to their complex network architecture and black box nature. They often make detection decisions based on high-dimensional feature maps, limiting their explainability.

To overcome these drawbacks, our contributions are summarized as follows:

$\bullet $ As the core idea of our contribution, we propose a Prototype-based Unified Framework for Deepfake Detection (PUDD) framework. Prototypes are clustered to learn representations for the upstream task, i.e., video classification. This robust representation allows deepfakes generated by unseen deepfake techniques to be returned unedited, maintaining their visual integrity and preserving their latent space representation.

$\bullet $ We propose integrating state-of-the-art techniques from sim-DNN \cite{simdnn} and xClass \cite{eclass1} for deepfake detection. Our approach includes a prototype learning layer that is easily trained and significantly enhances detection accuracy without necessitating the retraining of the entire framework. Additionally, it significantly reduces CO2 emissions, computational and power requirements compared to other large detection  and classification models making our approach significantly more environmentally friendly.

$\bullet$ We provide interpretability to understand the prototype-based classification as the degree to which a human can consistently predict the model’s output.

$\bullet $ We demonstrate the efficiency and effectiveness of our proposed methods by comparing them to state-of-the-art deepfake detection models across multi-modal data, i.e., deepfake images and videos.

\section{Related Works}
\subsection{Deepfake Generation}
Deepfake generation involves the use of deep learning techniques to create convincing image, audio and video hoaxes. There are several methods for creating deepfakes, but the most widely used methods are Variational Autoencoders (VAEs) \cite{faceswap, autodf}, Generative Adversarial Networks (GANs) \cite{gandf, gandf1, stylegan}, and diffusion models \cite{diffusion}. To generate a deepfake with VAEs, FaceSwap encodes both the source and target faces into the latent space using the trained encoder \cite{faceswap}. Then, it swaps the latent representations of the faces, effectively transferring the facial features of the target face onto the source face. Moreover, as a common used deepfake technique, style-based GAN (StyleGAN) \cite{stylegan} facilitates an automatically learned, unsupervised separation of high-level attributes, e.g., pose and identity when trained on human faces, and stochastic variation in the generated images, e.g., freckles and hair. It also allows for intuitive, scale-specific control of the synthesis.

\subsection{Deepfake Detection}
The recent literature \cite{polit, polit1, IJCNN, IJCNN2024} confirms the critical need for detecting deepfakes to protect the reputations and credibility of public figures, particularly politicians, who are vulnerable to manipulation and misinformation campaigns. Deepfakes have the potential to propagate false narratives and undermine trust in democratic processes. Robust detection methods are therefore essential to prevent the dissemination of deceptive content. By investing in deepfake detection technologies, we can mitigate the risks posed by malicious actors intent on exploiting digital media for political gain, thus safeguarding the integrity of public discourse

Reiss et al. introduce a state-of-the-art deepfake detection technique based on the concept of 'fact checking', adapted from fake news detection \cite{fdetect4}. This approach verifies that claimed facts (e.g., identity as Biden) align with observed media (e.g., is the face truly Biden's?), allowing differentiation between real and fake media. Similar with our upstream video classification task, Haliassos et al. propose self-supervised representation learning across visual and auditory modalities to capture factors such as facial movements, expression, and identity \cite{fdetect5}. These learned representations serve as targets predicted by the detector alongside the traditional binary forgery classification task.

\subsection{Prototype Learning}
As depicted in Figure 1, prototype-based deepfake detection methods \cite{dmitry, simdnn} calculate the local peaks of the density for each individual, essentially identifying the most representative data samples in each class from the training set as prototypes. 
\begin{figure}[t]
  \centering
   \includegraphics[width=8.2cm, height=2.8cm]{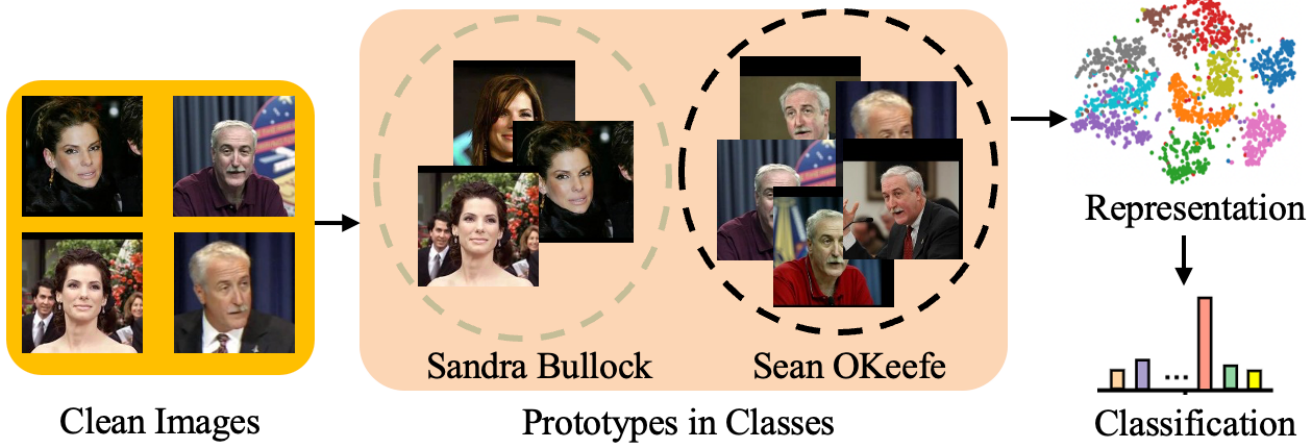}
   \caption{Prototype learning-based image classification with original images. }
   \label{fig:onecol}
\end{figure}

These methods then evaluate the similarity between new data samples and autonomously selected prototypes to classify images as either deepfake or original data samples. Bouter et al. simplify the complexity of working with spatio-temporal prototypes and enable their replacement to achieve greater interpretability \cite{prot1}. Aghasanli et al. calculate similarity scores (Euclidean distance in feature space) between an input image and all identified prototypes to derive rules for each specific sample \cite{dmitry}. However, a common limitation of prototype-based deepfake detection methods is the time-consuming nature of retraining detectors for new classes or individuals. 

\begin{figure*}[htbp!]
\centering
\includegraphics[width=16.4cm, height=4.5cm]{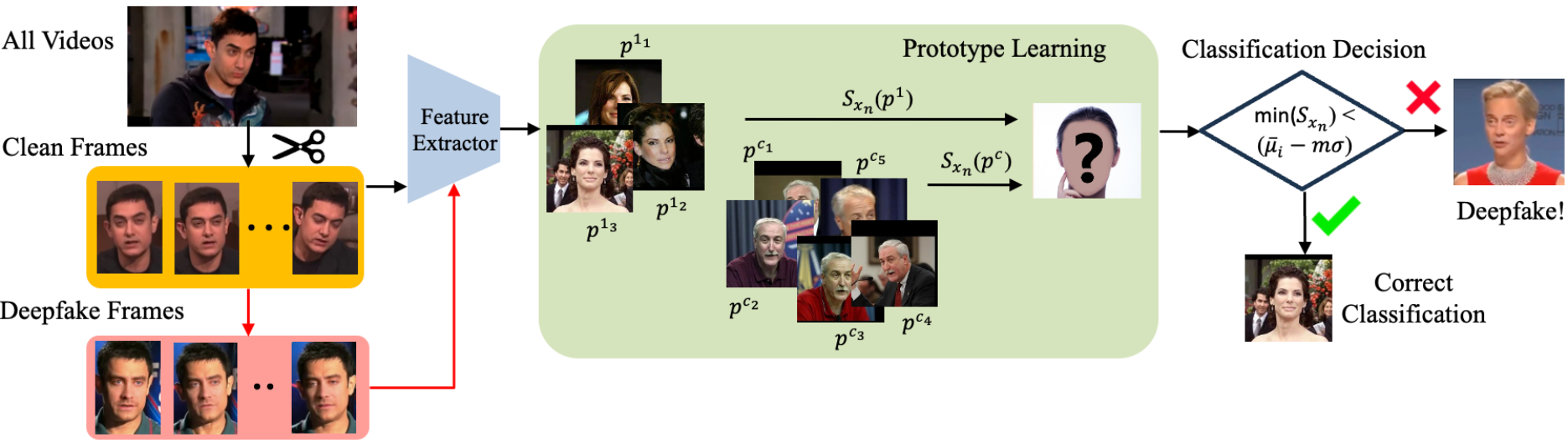}
\caption{The proposed prototype learning-based framework. We extract frames from raw videos and crop them into small patches. The \textit{red} lines only refer to the inference stage.}\centering
\end{figure*}
\section{Proposed Method}
Our proposed solution is built on a series of novel contributions that collectively form the deepfake detection framework. As illustrated in Figure 2, these innovations include the introduction of the Prototype Learning layer (\ref{3.3}) to cluster prototypes from input data and calculate their similarity to established prototypes. Additionally, the Classification layer (\ref{3.4}) is developed to classify images based on the estimated similarity scores obtained from the Prototype Learning layer.


\subsection{Pre-processing}
In our experiments, we evaluate our framework on two public datasets \cite{celeb, cifake}. Firstly, for the Celeb-DF dataset, we sample one frame every two seconds from the videos. These frames are then cropped to extract smaller patches containing only the face regions. In the training stage, we utilize all 59 celebrities available in the Celeb-DF dataset \cite{celeb}. However, different from conventional methods that rely on paired data, we only consider frames from original videos for training and frames from deepfake videos for inference. Secondly, for the CIFAKE dataset, no preprocessing of clean images is required prior to training.


\subsection{Feature Extraction}
The proposed PUDD extracts features from a pre-trained model for prototype learning and the upstream task, i.e., video classification, eliminating the need for fine-tuning. To achieve this, we choose DINOV2 \cite{dino} as the feature extractor due to its ability to effectively correct non-uniformities in images and its promising performance in image classification tasks. After feature extraction, the prototypes are calculated and learned from these features.

\subsection{Prototype Learning Layer} \label{3.3}
Given a training set $X=\{x_1,x_2,...,x_n\}$ of $n$ image samples with $C$ classes, we aim to learn the prototypes $P=\{p^{1_1},p^{1_2},...,p^{c_m}\}$ of original videos for video classification. For example, $p^{c_m}$ refers to the $m$-th prototype in the in the $c$-th class. Particularly, the most representative data samples in each class of the dataset are selected as prototypes. We show the prototype clustering result in Figure 3.

\begin{figure}[htbp!]
\centering
\includegraphics[width=8.3cm, height=7.3cm]{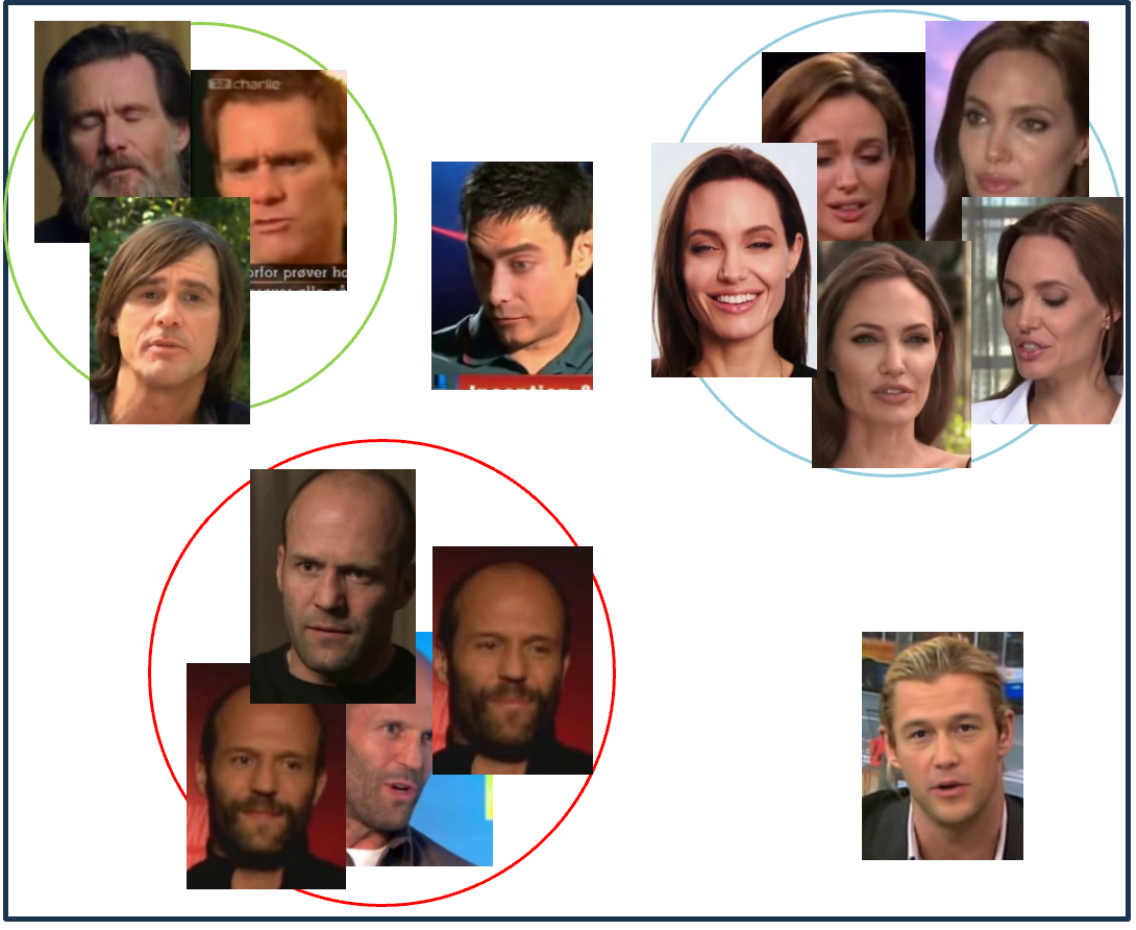}
\caption{Prototype clustering visualizations. }\centering
\end{figure}

In Figure 3, it is depicted that the prototypes from original videos (ID13, Id23, and Id24) are clustered, while two outliers, i.e., deepfake videos are kept away from these clusters. Specifically, even though the hair style, presence of a mustache, and apparent age vary across data samples within ID13, the prototypes are clustered effectively to enable successful classification of the celebrity. This indicates that the proposed PUDD framework can capture and leverage subtle yet discriminative features to distinguish between different individuals, even amidst significant variations in appearance. These prototypes facilitate a reasoning process based on the similarity (proximity in feature space) between a data sample and a prototype. In this case, prototypes are identified as the local density peaks \cite{simdnn}, essentially the most representative samples from the training set. Therefore, only a limited number of samples from the training dataset are chosen as prototypes, ensuring the system’s efficiency and compatibility with a broad range of devices. We define $N_c$ and $M_c$ are the numbers of samples and prototypes in the $c$-th class, respectively. 

As the core idea of our contribution, the Prototype Learning layer serves to cluster prototypes and calculate the similarity and for each image associated prototype. Inspired from a specific Cauchy equation in \cite{simdnn}, we define a similarity score between $n$-th data $x_n$ and prototypes in $c$-th class by using Euclidean distance to identify how closely new data aligns with known data patterns drawn from the extracted features:
\begin{equation}
S_{x_n}(p^c)=\frac{\sum_{m=1}^{M_c}\left\|x_n- p^{c_m}\right\|^2}{1+\frac{\|x-\mu\|^2}{\|\sigma\|^2}}
\end{equation}
where $\mu$ and $\sigma$ are the mean and variance of data samples, respectively. This step evaluates the proximity of data samples within the feature space, utilizing Euclidean distance as metrics. After calculating all the similarity scores to prototypes in all classes, we compare the minimum similarity score against the mean and variance of data by using m-$\sigma$ rule.

\subsection{Classification Layer} \label{3.4}
The proposed classification layer makes the decision on whether an input belongs to an existing class or a deepfake. The m-$\sigma$ rule is applied to detect potential attacks, which can be depicted through an inequality condition:
\begin{equation}
\begin{aligned}
\text { IF } & \min \left(S_{x_n}\right)>(\bar{\mu}-m \sigma) \\
\text { THEN } & x_n \in \text { Potential deepfake video or image } \\
\text { ELSE } & x_n \in \text { Classification label }
\end{aligned}
\end{equation}
where $\bar{\mu}$ refers to the recursive mean of data samples. If this condition is met, it suggests that the system has recognized a divergence or a new data concept, distinct from the established data patterns used to generate the prototypes. If not, it indicates no significant change in the data concept, allowing the algorithm to continue with its standard classification process. This mechanism enables PUDD to adaptively respond to new data and effectively identify potential deepfake detection.

\section{Experiments}
\subsection{Data and Deepfakes}
\subsubsection{Celeb-DF} 
In the Celeb-DF dataset \cite{celeb}, there are 590 real videos featuring 59 celebrities of diverse genders, ages, and ethnic groups, collected from publicly available sources such as YouTube. Additionally, the dataset includes 5,639 deepfake videos generated using improved synthesis methods, including color mismatch, inaccurate face masks, and temporal flickering. Consequently, the overall visual quality of the synthesized deepfake videos in Celeb-DF is significantly enhanced compared to existing datasets, with notably fewer visual artifacts.

In this work, we use all 590 real videos and 5,639 deepfake videos for the training and inference stages, respectively. We extract one frame per every two seconds of the videos and then crop these frames into smaller patches containing only the face regions. Specifically, the training stage comprises 11,723 cropped frames from the original videos, while the inference set consists of 60,847 cropped frames from the deepfake videos.

\subsubsection{CIFAR-10}
Different from Celeb-DF, CIFAKE \cite{cifake} is designed to encompass non-human classes such as birds, cars, and ships. The dataset comprises 60,000 synthetically-generated images and an equal number of real images collected from CIFAR-10 \cite{cifar10}. The synthetic images are generated using a fine-tuned Stable Diffusion Model \cite{stablefu}.

In our study, we train the model using prototypes learned from 50,000 original images in the training set. Subsequently, we evaluate the proposed method using 50,000 deepfake images from the inference set.

\begin{table*}[htbp!]
\centering
\caption{Deepfake video detection comparison on the Celeb-DF dataset. Para. and Acc are trainable parameters and detection accuracy, respectively.}
\begin{tabular}{c|ccc|cc|c}
\hline
 & \multicolumn{3}{c|}{Algorithm $\&$ Network}  &\multicolumn{2}{c|}{Computational Cost}  & Acc ($\%$)\\ 
\hline
Method &  Pre-training & Prototype & Backbone &Para.& Training Time (s)  &\\ 
\hline 
MPC-CA \cite{proadd} &\checkmark &\checkmark & BERT + MLP & 6.4 M & 399.5 &79.5  \\
Sim-DNN \cite{simdnn} &\checkmark & \checkmark & VGG16 + DNN & {\bfseries 2.8 M} & 109.8&88.4 \\
IPD \cite{dmitry} &\checkmark &\checkmark & ViT-L-32 + MLP Head & 3.7 M & 254.7 & 89.2  \\	
ProtoExplorer \cite{prot1} & \xmark & \checkmark & DPNet & 3.8 M & 258.0 & 92.5  \\
\hline 
NoiseDF \cite{fdetect3} &\checkmark &\xmark & RIDNet + Attention & 9.9 M& 278.6 & 70.1  \\	
FTCN \cite{FTCN} &\xmark &\xmark & FTCN & 26.6 M & 7482.6 & 86.9  \\
MMtrace \cite{fdetect2} &\xmark & \xmark & MLP & 4.7 M & 209.5 & 92.9 \\
\hline
 \textit{PUDD} & \checkmark & \checkmark & DINOV2 + xDNN &  7.6 M & {\bfseries 2.7} & {\bfseries 95.1}  \\
 \hline 
\end{tabular}
\end{table*}
\subsection{Competitors and Implementation}
The proposed method is evaluated and compared to state-of-the-art competitor models. We reproduce three state-of-the-art deepfake detection techniques \cite{fdetect2, fdetect3, FTCN}, utilizing the best-reported implementations available in the literature. For example, Aghasanli et al. achieve superior results by fine-tuning only the multilayer perceptron (MLP) head in the original Vision Transformer (ViT) \cite{dmitry}. Therefore, we fine-tune this model with our dataset to serve as a competitor in our comparison experiments. Secondly, we reproduce four prototype learning methods, including both for deepfake detection \cite{prot1, dmitry} and adversarial attack detection \cite{simdnn, proadd}, i.e., similar to deepfakes. 

In this paper, the proposed prototype learning is implemented on the detector and further studies on feature extractor are out of scope of this paper. In the comparison experiments (\ref{celeb} $\&$ \ref{cifake}), we exploit DINOV2 \cite{dino} as the feature extractor. All the experiments are run on Tesla V100 GPUs. 

\section{Results}
\subsection{Celeb-DF} \label{celeb}
The proposed PUDD is evaluated on deepfake detection task over the Celeb-DF dataset \cite{celeb}. Table 1 shows the results, each of them is the average of 60,847 deepfake frames.

From Table 1, it can be observed that: (1) In all the evaluated models, the proposed PUDD achieves \textit{95.1$\%$} for deepfake video detection, which offers the best effectiveness. (2) PUDD demonstrates state-of-the-art efficiency in the training stage compared to state-of-the-art models due to its rapid calculation of simple prototypes. This feature enables swift retraining for unseen celebrities, making it highly practical for real-world applications.
\subsection{CIFAKE}  \label{cifake}
We compare the deepfake image detection performance over the CIFAKE dataset \cite{cifake}. The results are presented in Table 2, each result is average of 50,000 deepfake images.

\begin{table}[htbp!]
\centering
\caption{Deepfake image detection comparison on the CIFAKE dataset.}
\begin{tabular}{c|c|c}
\hline
Method &  Training Time (s)  & Acc ($\%$)\\ 
\hline 
MPC-CA \cite{proadd} & 483.2&79.5  \\
Sim-DNN \cite{simdnn} & 142.3&88.4 \\
IPD \cite{dmitry} & 249.9 & 89.2  \\	
ProtoExplorer \cite{prot1} & 261.7 & 92.5  \\
\hline 
NoiseDF \cite{fdetect3} & 300.6 & 70.1  \\	
FTCN \cite{FTCN} & 8358.3 & 86.9  \\
MMtrace \cite{fdetect2} & 252.7 & 92.9 \\
\hline
 \textit{PUDD} & {\bfseries 2.8} & {\bfseries 94.6}  \\
 \hline 
\end{tabular}
\end{table}

From Table 2, the proposed PUDD outperforms the state-of-the-art models \cite{proadd, simdnn, dmitry, prot1, fdetect3, FTCN, fdetect2} on both accuracy and training time. there are three main differences between the Celeb-DF and CIFAKE datasets. Firstly, Celeb-DF comprises video data, whereas CIFAKE consists of image data. Secondly, the deepfake generation techniques used in these datasets differ, with Celeb-DF employing improved generation techniques and CIFAKE utilizing the Stable Diffusion Model. Thirdly, while Celeb-DF only includes human classes, CIFAKE encompasses 10 non-human classes such as cars, birds, ships, and cats. Therefore, the robust deepfake detection performance observed across these two datasets validates the effectiveness of PUDD across diverse scenarios.
\subsection{Visualization}
In this section, we make some visualizations to confirm the effectiveness of the proposed PUDD framework. Firstly, Figure 4 illustrates different similarity scores when different deepfakes generated from original videos are considered.

\begin{figure}[htbp!]
\centering
\includegraphics[width=8.2cm, height=4.8cm]{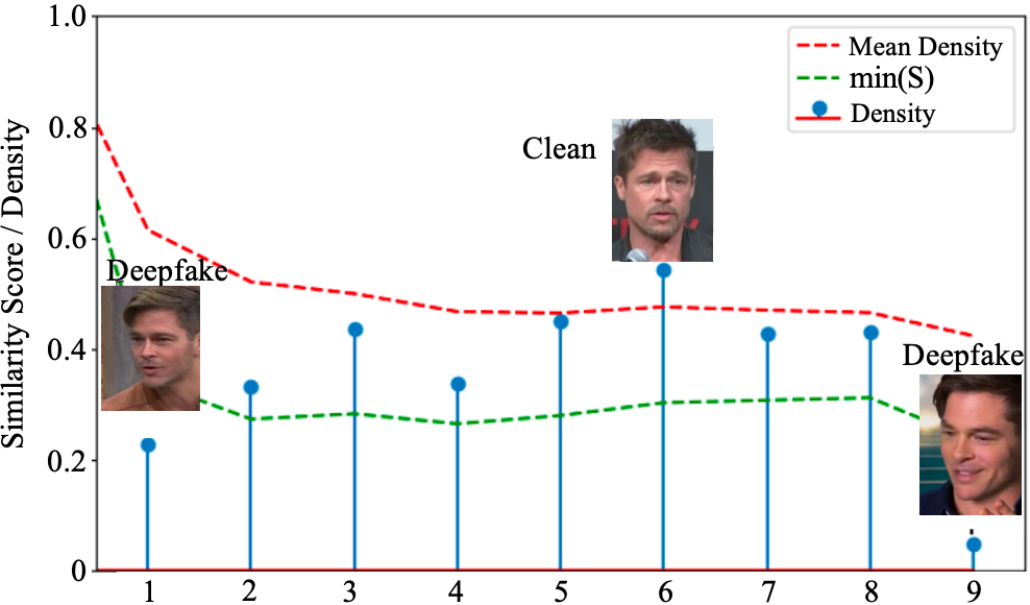}
\caption{Similarity/Density score drop in deepfake videos.}\centering
\end{figure}

From Figure 4, it is apparent that for each incoming sample, we can calculate and plot their density score relative to the existing prototypes, the mean density of our prototypes, and the minimum of similarity score. By considering these values, we can visually discern that a clean image will exhibit a density score higher than the minimum of similarity score, whereas DeepFaked images will be flagged as abnormal and will have a value below the minimum of similarity score.

Secondly, we present some qualitative result in Figure 5 to show the effectiveness of PUDD. The detection results of PUDD and MMtrace are denoted by green/red and black, respectively.

\begin{figure}[htbp!]
\centering
\includegraphics[width=8.2cm, height=6.2cm]{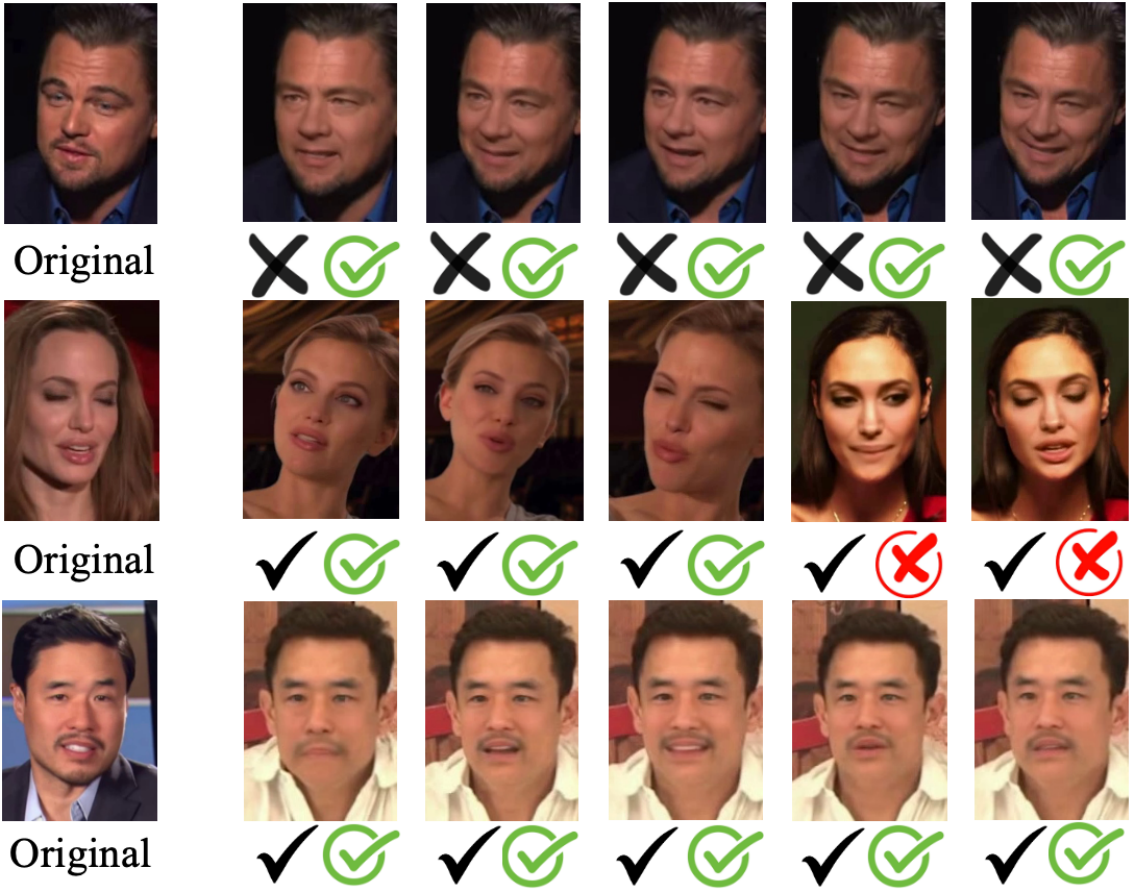}
\caption{Challenging deepfakes in Celeb-DF. Black and green/red marks refer to the detection prediction from the MMtrace and PUDD, respectively.}\centering
\end{figure}

As qualitative analysis, Figure 5 presents the deepfake detection results by using MMtrace and PUDD. We can observe that: (1) Both PUDD and MMtrace successfully detect the third celebrity as it is relatively easy to distinguish; (2) PUDD outperforms MMtrace in detecting the first celebrity by classifying them as an unseen class, leading to a more accurate decision; (3) PUDD fails to detect the second video of the second celebrity. This failure may be due to the celebrity frequently closing their eyes throughout the majority of the video, making prototype recognition more challenging.
\subsection{Image Classification}
As aforementioned, we estimate the prototypes for image classification as the upstream task, demonstrating promising performance of PUDD in both image classification and deepfake detection tasks during inference. In this experiment, we compare the image classification accuracy over original videos and images in the Celeb-DF \cite{celeb} and CIFAKE datasets \cite{cifake}, respectively. The results are presented in Table 3.

\begin{table}[htbp!]
\centering
\caption{Image classification comparison on Celeb-DF and CIFAKE.}
\begin{tabular}{c|c|c}
\hline
Method &  Celeb-DF  & CIFAKE\\ 
\hline 
MPC-CA \cite{proadd} & 79.2 & 84.1  \\
IPD \cite{dmitry} & 87.4 &  87.9 \\	
ProtoExplorer \cite{prot1} & 92.0 & 94.6  \\
\hline 
NoiseDF \cite{fdetect3} & 90.1 & 94.7  \\	
FTCN \cite{FTCN} & 92.3 & 93.6  \\
MMtrace \cite{fdetect2} & 92.5 & 93.9 \\
\hline
 \textit{PUDD} & {\bfseries 92.7} & {\bfseries 96.4}  \\
 \hline 
\end{tabular}
\end{table}

It can be observed from Table 3 that PUDD achieves best image classification accuracy on both datasets, i.e., 92.7$\%$ and 96.4$\%$, respectively. These results affirm the promising performance of PUDD across both tasks.
\subsection{Interpretability}
As aforementioned, the proposed PUDD learns prototypes from the data samples to provide interpretability. We calculate the similarity score (as described in Eq. 1) between an input image and all identified prototypes, so, we were able to extract a rule-based linguistic representation for each specific sample to explain the model’s behavior as described:

\begin{figure}[htbp!]
\centering
\includegraphics[width=8.2cm, height=1.6cm]{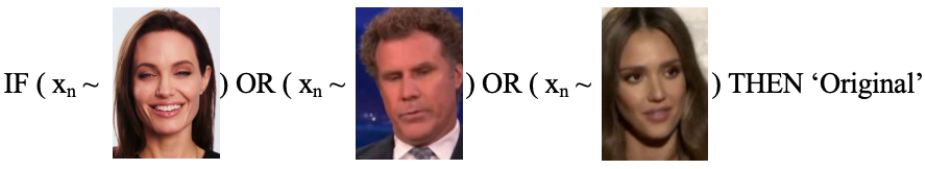}
\caption{Linguistic rule-based representation of the prototypes for PUDD Interpretability on ’Original’ class of Celeb-DF with top 3 closest prototypes on the feature space.}\centering
\end{figure}

\subsection{Environmental Impact}
As aforementioned, PUDD only requires a limited number of parameters for retraining due to efficient prototype learning. We report the potential carbon emission of retraining a PUDD in Table 4. All models are trained on a single V100 GPU with a power consumption of 300 W.

\begin{table}[htbp!]
\centering
\caption{Carbon footprint of reproducing models. $\mathrm{tCO}_2 \mathrm{eq}$ refers to the tonnes of $\mathrm{CO}_2$ equivalent.}
\begin{tabular}{cccccc}
\hline
Method  & Total Power Consumption & $\mathrm{tCO}_2 \mathrm{eq}$\\ 
\hline 
NoiseDF \cite{fdetect3} & 23.2 kWh & 1.2 $\times 10^{-2}$  \\	
FTCN \cite{FTCN} & 624.4 kWh & 0.3  \\
MMtrace \cite{fdetect2} & 17.5 kWh & 8.7$\times 10^{-3}$ \\
\hline
 \textit{PUDD}  & 0.2 Wh & $10^{-7}$ \\
 \hline 
\end{tabular}
\end{table}

For comparison, retraining a MMtrace or PUDD would require 17.5 kWh and 0.2 Wh, respectively, if run in the same data center. This is $10^{5}$ more carbon emission.

\section{Discussion and Conclusion}
The advantages of our proposed method are listed bellow:

1. In the training stage, our approach only require the access to original data. Therefore, different from conventional deepfake detection methods, we do not rely on paired training data, which includes both original and deepfake samples. This characteristic of our method streamlines the training process and eliminates the need for paired samples, simplifying the data collection and labeling process.

2. The proposed PUDD exploits prototype information derived from original data, thereby rendering it agnostic to the specific deepfake generation techniques and models present in the inference data. Consequently, it exhibits the capability to effectively detect unseen deepfakes generated using different techniques and models than those encountered during the training stage. The experimental results further confirm the effectiveness of PUDD.

3. PUDD can be easily implemented with various feature extractors to detect deepfakes across diverse data modalities, including video and image. Moreover, PUDD offers flexibility for researchers to select a suitable feature extractor tailored to the specific characteristics of their target class.

4. The rapid retraining capability of PUDD, taking only 2.7 seconds, significantly accelerates its adoption in new domains compared to conventional deepfake detection methods. This speed makes PUDD highly feasible for potential real-world applications, enhancing its practicality and versatility.

5. The PUDD framework exploits image classification as the upstream task, enabling it to achieve promising performance in image classification despite being primarily designed for deepfake detection and trained on a deepfake dataset. This demonstrates the adaptability and robustness of PUDD across various tasks and datasets.

6. Prototype learning aids in understanding prototype-based classification by quantifying the extent to which a human can reliably predict the model's output.

7. Due to efficient prototype clustering and simplified calculations, PUDD requires $10^5$ times less carbon emission than the state-of-the-art model, making it much more environmentally friendly.

Overall, we have proposed a Prototype-based
Unified Framework for Deepfake Detection (PUDD) framework for deepfake video and image detection, offering an effective alternative to conventional competitors. Different from these conventional methods, we learned most representative prototypes in classes to efficiently detect deepfake samples and provide interpretability. Our evaluation with multi-modal datasets has demonstrated the robust performance of the proposed method on both deepfake images and videos. Additionally, PUDD required only 2.7 seconds on new data, making it feasible for potential real-world applications.

\section*{Acknowledgment}
This work is supported by ELSA – European Lighthouse on Secure and Safe AI funded by the European Union under grant agreement No. 101070617. Views and opinions expressed are however those of the author(s) only and do not necessarily reflect those of the European Union or European Commission. Neither the European Union nor the European Commission can be held responsible.
\bibliographystyle{ieee_fullname}
\bibliography{PaperForReview}

\begin{thebibliography}{10}\itemsep=-1pt

\bibitem{deepfake}
A. Aghasanli, D. Kangin, and P. Angelov.
\newblock {Interpretable-through-prototypes deepfake detection for diffusion models}.
\newblock {\em Proceedings of IEEE/CVF Conference on Computer Vision and Pattern Recognition (CVPR)}, 2023.

\bibitem{dmitry}
A. Aghasanli, D. Kangin, and P. Angelov.
\newblock {Interpretable-through-prototypes deepfake detection for diffusion models}.
\newblock {\em Proceedings of IEEE/CVF International Conference on Computer Vision (ICCV)}, 2023.

\bibitem{eclass1}
P. Angelov and E. Soares.
\newblock {Detecting and learning from unknown by extremely weak supervision: exploratory classifier (xclass)}.
\newblock {\em Neural Computing and Applications}, 33:15145–15157, 2021.

\bibitem{polit}
M. Appel and F. Prietzel.
\newblock {The detection of political deepfakes}.
\newblock {\em Journal of Computer-Mediated Communication}, 27(4):1 -- 13, 2022.

\bibitem{cifake}
J.~J. Bird and A. Lcifi.
\newblock {CIFAKE: image classification and explainable identification of AI-generated synthetic images}.
\newblock {\em IEEE Access}, page~99, 2024.

\bibitem{prot1}
M.~L. Bouter, J.~L. Pardo, Z. Geradts, and M. Worring.
\newblock {ProtoExplorer: interpretable forensic analysis of deepfake videos using prototype exploration and refinement}.
\newblock {\em arXiv preprint arXiv:2309.11155}, 2023.

\bibitem{news}
S. Burgess.
\newblock {Ukraine war: deepfake video of Zelenskyy telling Ukrainians to 'lay down arms' debunked}.
\newblock {\em Sky News}, 2023.

\bibitem{fdetect}
A. Chintha, A. Rao, S. Sohrawardi, K. Bhatt, M. Wright, and R. Ptucha.
\newblock {Leveraging edges and optical flow on faces for deepfake detection}.
\newblock {\em Proceedings of IEEE International Joint Conference on Biometrics (IJCB)}, 2020.

\bibitem{Xception}
F. Chollet.
\newblock {Xception: deep learning with depthwise separable convolutions}.
\newblock {\em Proceedings of IEEE/CVF Conference on Computer Vision and Pattern Recognition (CVPR)}, 2017.

\bibitem{vit}
A. Dosovitskiy, L. Beyer, A. Kolesnikov, D. Weissenborn, X. Zhai, T. Unterthiner, M. Dehghani, M. Minderer, G. Heigold, S. Gelly, J. Uszkoreit, and N. Houlsby.
\newblock {An image is worth 16x16 words: Transformers for image recognition at scale}.
\newblock {\em Proceedings of International Conference on Learning Representations (ICLR)}, 2021.

\bibitem{gan}
I. Goodfellow, J. Pouget-Abadie, M. Mirza, B. Xu, D. Warde-Farley, S. Ozair, A. Courville, and Y. Bengio.
\newblock {Generative adversarial nets}.
\newblock {\em Conference on Neural Information Processing Systems (NeurIPS)}, 2014.

\bibitem{faceswap}
Y. Guo, W. He, J. Zhu, and C. Li.
\newblock {A light autoencoder networks for face swapping}.
\newblock {\em Proceedings of International Conference on Computer Science and Artificial Intelligence (ICCSAI)}, 2018.

\bibitem{nst}
D. Gutiérrez and M. Mendoza.
\newblock {Bimodal neural style transfer for image generation based on text prompts}.
\newblock {\em Proceedings of International Conference on Human-Computer Interaction}, 2023.

\bibitem{fdetect5}
A. Haliassos, R. Mira, S. Petridis, and M. Pantic.
\newblock {Leveraging real talking faces via self-supervision for robust forgery detection}.
\newblock {\em Proceedings of IEEE/CVF Conference on Computer Vision and Pattern Recognition (CVPR)}, 2022.

\bibitem{diffusion}
J. Ho, A. Jain, and P. Abbeel.
\newblock {Denoising diffusion probabilistic models}.
\newblock {\em Advances in neural information processing systems}, 33:6840--6851, 2020.

\bibitem{fake}
Y. Hou, Q. Guo, Y. Huang, X. Xie, L. Ma, and J. Zhao.
\newblock {Evading deepFake detectors via adversarial statistical consistency}.
\newblock {\em Proceedings of IEEE/CVF Conference on Computer Vision and Pattern Recognition (CVPR)}, 2023.

\bibitem{polit1}
J. Ice.
\newblock {Defamatory political deepfakes and the first amendment}.
\newblock {\em Case Western Reserve Law Review}, 70(2):417 -- 455, 2019.

\bibitem{new1}
R. Jones.
\newblock {Real person or deepfake? can You tell?}
\newblock {\em University of Waterloo}, 2023.

\bibitem{detectc1}
E. Josephs, C. Fosco, and A. Oliva.
\newblock {Artifact magnification on deepfake videos increases human detection and subjective confidence}.
\newblock {\em Journal of Vision}, 23:5327, 2023.

\bibitem{stylegan}
T. Karras, S. Laine, and T. Aila.
\newblock {A style-based generator architecture for generative adversarial networks}.
\newblock {\em Proceedings of IEEE/CVF Conference on Computer Vision and Pattern Recognition (CVPR)}, 2019.

\bibitem{detectc2}
A. Khormali and J.-S. Yuan.
\newblock {Self-supervised graph Transformer for deepfake detection}.
\newblock {\em arXiv preprint arXiv:2307.15019}, 2023.

\bibitem{cifar10}
A. Krizhevsky.
\newblock Learning multiple layers of features from tiny images.
\newblock {\em Master's thesis}, 2009.

\bibitem{gandf1}
C.-H. Lee, Z. Liu, L. Wu, and P. Luo.
\newblock {MaskGAN: towards diverse and interactive facial image manipulation}.
\newblock {\em Proceedings of IEEE/CVF Conference on Computer Vision and Pattern Recognition (CVPR)}, 2020.

\bibitem{IJCNN}
Y. Li, P. Angelov, and N. Suri.
\newblock {Domain generalization and feature fusion for cross-domain imperceptible adversarial attack detection}.
\newblock {\em Proceedings of the International Joint Conference on Neural Networks (IJCNN)}, 2023.

\bibitem{IJCNN2024}
Y. Li, P. Angelov, and N. Suri.
\newblock {Rethinking self-supervised learning for cross-domain adversarial sample recovery}.
\newblock {\em Proceedings of International Joint Conference on Neural Networks (IJCNN)}, 2024.

\bibitem{TAI}
Y. Li, Y. Sun, K. Horoshenkov, and S.~M. Naqvi.
\newblock {Domain adaptation and autoencoder based unsupervised speech enhancement}.
\newblock {\em IEEE Transactions on Artificial Intelligence}, 3(1):43 -- 52, 2021.

\bibitem{celeb}
Y. Li, X. Yang, P. Sun, H. Qi, and S. Lyu.
\newblock {Celeb-DF: a large-scale challenging dataset for deepfake forensics}.
\newblock {\em Proceedings of IEEE/CVF Conference on Computer Vision and Pattern Recognition (CVPR)}, 2020.

\bibitem{fake1}
K. Narayan, H. Agarwal, K. Thakral, S. Mittal, M. Vatsa, and R. Singh.
\newblock {DF-Platter: multi-face heterogeneous deepfake dataset}.
\newblock {\em Proceedings of IEEE/CVF Conference on Computer Vision and Pattern Recognition (CVPR)}, 2023.

\bibitem{dino}
M. Oquab, T. Darcet, T. Moutakanni, H. Vo, M. Szafraniec, V. Khalidov, P. Fernandez, D. Haziza, F. Massa, A. El-Nouby, M. Assran, N. Ballas, W. Galuba, R. Howes, P.-Y. Huang, S.-W. Li, I. Misra, M. Rabbat, V. Sharma, G. Synnaeve, H. Xu, H. Jegou, J. Mairal, P. Labatut, A. Joulin, and P. Bojanowski.
\newblock Dinov2: learning robust visual features without supervision.
\newblock {\em arXiv preprint arXiv: 2304.07193}, 2023.

\bibitem{fdetect2}
M.~A. Raza and K. Malik.
\newblock {Multimodaltrace: deepfake detection using audiovisual representation learning}.
\newblock {\em Proceedings of IEEE/CVF Conference on Computer Vision and Pattern Recognition (CVPR)}, 2023.

\bibitem{fdetect4}
T. Reiss, B. Cavia, and Y. Hoshen.
\newblock {Detecting deepfakes without seeing any}.
\newblock {\em arXiv preprint arXiv:2311.01458}, 2023.

\bibitem{stablefu}
R. Rombach, A. Blattmann, D. Lorenz, P. Esser, and B. Ommer.
\newblock {High-resolution image synthesis with latent diffusion models}.
\newblock {\em arXiv preprint arXiv:2112.10752}, 2021.

\bibitem{gandf}
P. Sharma, M. Kumar, and H.~K. Sharma.
\newblock {A GAN-based model of deepfake detection in social media}.
\newblock {\em Procedia Computer Science}, 218:2153--2162, 2023.

\bibitem{simdnn}
E. Soares, P. Angelov, and N. Suri.
\newblock {Similarity-based deep neural network to detect imperceptible adversarial attacks}.
\newblock {\em Proceedings of IEEE Symposium Series on Computational Intelligence (SSCI)}, 2022.

\bibitem{autodf}
D.-C. Stanciu and B. Ionescu.
\newblock {Autoencoder-based data augmentation for deepfake detection}.
\newblock {\em Proceedings of Annual ACM International Conference on Multimedia Retrieval (ICMR)}, 2023.

\bibitem{fdetect3}
T. Wang and K. Chow.
\newblock Noise based deepfake detection via multi-head relative-interaction.
\newblock {\em Proceedings of the Association for the Advancement of Artificial Intelligence (AAAI)}, 2023.

\bibitem{aedeep}
M. Zendrana and A. Rusiecki.
\newblock {Swapping face images with generative neural networks for Deepfake technology – experimental study}.
\newblock {\em Proceedings of International Conference on Knowledge-Based and Intelligent Information and Engineering Systems}, 2021.

\bibitem{proadd}
F. Zhang, S. Tian, L. Yu, and Q. Yang.
\newblock {Multi-channels prototype contrastive learning with condition adversarial attacks for few-shot event detection}.
\newblock {\em Neural Processing Letters}, 56(31):30 -- 31, 2024.

\bibitem{fdetect1}
C. Zhao, C. Wang, G. Hu, H. Chen, C. Liu, and J. Tang.
\newblock {ISTVT: interpretable spatial-temporal video Transformer for deepfake detection}.
\newblock {\em IEEE Transactions on Information Forensics and Security}, 18:1335 -- 1348, 2023.

\bibitem{FTCN}
Y. Zheng, J. Bao, D. Chen, M. Zeng, and F. Wen.
\newblock {Exploring temporal coherence for more general video face forgery detection}.
\newblock {\em Proceedings of IEEE/CVF International Conference on Computer Vision (ICCV)}, 2021.

\end{thebibliography}

\end{document}